\documentclass[10pt,journal,compsoc]{IEEEtran}

\usepackage{amsmath, amssymb, amsfonts, mathtools, bm}
\usepackage{amsthm}
\usepackage{nicefrac}
\usepackage{latexsym}
\usepackage{graphicx}
\usepackage{subfigure}
\usepackage{wrapfig}
\usepackage{float}
\usepackage{adjustbox}
\usepackage{stfloats}
\usepackage{booktabs}
\usepackage{multirow}
\usepackage{makecell}
\usepackage{colortbl}
\usepackage{threeparttable}
\usepackage{arydshln}
\usepackage{algorithm}
\usepackage{caption}
\usepackage{textcomp}
\usepackage{enumitem}
\usepackage{pifont}
\usepackage{bbding}
\usepackage{microtype}
\usepackage{setspace}
\usepackage{inconsolata}
\usepackage[T1]{fontenc}
\usepackage[utf8]{inputenc}
\usepackage[numbers]{natbib}
\usepackage{url}
\usepackage[table,svgnames]{xcolor}
\usepackage{hyperref}
\usepackage[capitalize,noabbrev]{cleveref}

\usepackage{algorithm}
\usepackage{algorithmicx}
\usepackage{algpseudocode}

\definecolor{papergreen}{cmyk}{0.91,0,0.88,0.12}
\definecolor{cpgcolor}{HTML}{DAE8FC}
\definecolor{lightblue}{HTML}{DAE8FC}

\hypersetup{
    colorlinks=true,
    linkcolor=blue,
    filecolor=magenta,      
    urlcolor=cyan,
    pdfborderstyle={/S/U/W 1} 
}
\begin{document}


\title{Mask IPL: Noise-Free Intrinsic Position Learning via Computation Graph Clipping for Event-Based Spike-Driven Tracking}

\author{Yimeng Shan,
         Malu~Zhang \\
\thanks{Y. Shan and M. Zhang are with the University of Electronic Science and Technology of China, Chengdu 610054, China, and also with the Shenzhen Loop Area Institute, Shenzhen 518038, China.}
}

\markboth{}%
{Shell \MakeLowercase{\textit{et al.}}: Bare Demo of IEEEtran.cls for Computer Society Journals}

\IEEEtitleabstractindextext{
\begin{abstract}
Spiking neural networks (SNNs) match the event-driven nature of event cameras and naturally extract spatiotemporal features. These properties have given rise to a series of recent studies on event-based tracking with SNNs. Because Intrinsic Position Learning (IPL) acquires stronger position information without introducing any parameters, it has become one of the mainstream approaches to acquiring position information in event-based spike-driven tracking. However, the mechanism behind its effectiveness still lacks a systematic theoretical analysis. Moreover, our analysis shows that IPL introduces additional noise in both forward and backward propagation. The former increases the inference error, and the latter prevents the parameters from converging to a better solution. This paper presents a systematic analysis of IPL and reveals that its effectiveness stems from the synergy between IPL and multi-stage convolution. The zero blocks of the joint tensor are equivalent to zero padding for convolution, and the resulting boundary effect spreads layer by layer through the multi-stage convolution. Every parameter update is therefore driven by a gradient that perceives the relative displacement between the template and search frames. A positional encoding added after the convolutional stage cannot provide this information. We further propose a simple Computation Graph Clipping method. It applies a validity mask determined by the layout to the operations of every layer, so that the invalid regions become equivalent to zero padding in both forward and backward propagation. This eliminates the above noise without introducing any parameters and makes the actual gradient coincide with the ideal gradient. We name the improved method Mask IPL. Without increasing the number of parameters or the computational cost, Mask IPL improves the AUC of the Tiny-scale tracker on FE108, FELT, and VisEvent and consistently improves the Base-scale tracker as well.

\end{abstract}

}

\maketitle
\IEEEdisplaynontitleabstractindextext
\IEEEpeerreviewmaketitle
\section{Introduction}




Event cameras capture visual information asynchronously and output sparse event streams, properties that match the spike-driven computation of spiking neural networks (SNNs). Combining SNNs with event cameras for low-power, high-temporal-resolution vision applications is therefore an important research direction~\cite{gallego2020event}. Among these applications, event-based single object tracking (SOT), a fundamental vision task, has drawn wide attention~\cite{wang2023visevent,zhang2021object} because it requires modeling the temporal information in the event stream and extracting the target's spatial features.

Early SNN-based trackers for event cameras usually contain floating-point multiplications and therefore cannot fully exploit the advantages of SNNs in event-driven computation and low power consumption~\cite{zhang2022spiking,zhang2025spiking}. To the best of our knowledge, SDTrack~\cite{shan2026sdtrack}, proposed by Shan et al., is the first event-based SOT framework built entirely from SNNs. Its results show that a network built entirely from SNNs reaches or even surpasses the tracking accuracy of conventional artificial neural networks (ANNs) while considerably reducing the inference energy. The success of SDTrack has made event-based tracking a common benchmark for evaluating SNN vision models and related techniques. Since then, many studies have explored network architectures~\cite{wang2026bipolar}, model compression~\cite{wei2026tp}, optimizers~\cite{zhouadas}, plug-and-play attention modules~\cite{shanspiking}, and loss functions~\cite{yang2026fully}, and have validated their effectiveness on event-based tracking.

Many of these methods acquire position information with Intrinsic Position Learning (IPL) rather than with the explicit positional encoding of the conventional SOT pipeline~\cite{dosovitskiy2020image,ye2022joint,chen2022backbone}. Existing results show that IPL generally outperforms explicit positional encoding in tracking performance. However, the mechanism behind the effectiveness of IPL still lacks a systematic analysis. We further find that IPL introduces additional feature noise in forward propagation and additional gradient noise in backward propagation. The former increases the inference error, and the latter disturbs parameter optimization and prevents the model from converging to a better solution.

To address these issues, this paper presents a in-depth analysis of IPL. We find that the effectiveness of IPL stems from its synergy with multi-stage convolution. The zero blocks of the joint tensor are equivalent to zero padding for convolution, and the resulting boundary effect spreads layer by layer through the multi-stage convolution. Every parameter update is therefore driven by a gradient that perceives the relative displacement between the template and search frames. Such cross-frame relative position information is absent from a positional encoding added after the convolutional stage. This finding explains why IPL outperforms explicit positional encoding and offers a new perspective on IPL-based trackers. On this basis, we propose a simple Computation Graph Clipping method. It applies a validity mask determined by the layout to the operations of every layer, so that the invalid regions become equivalent to zero padding in both forward and backward propagation. This eliminates the noise without introducing any parameters and makes the actual gradient coincide with the ideal gradient. We name the improved method Mask IPL. Without introducing any additional parameters or computational cost, Mask IPL improves the AUC of the Tiny-scale tracker on FE108, FELT, and VisEvent and also yields consistent improvements on the Base-scale tracker. These results indicate that efficiently modeling the relative position between the template and search frames is likely one of the key factors for improving event-based SOT. They also point to a new direction for the design of IPL-based trackers.

\section{Related Work}




Early attempts to introduce SNNs into event-based trackers were built on Siamese architectures. Zhang et al. proposed STNet~\cite{zhang2022spiking}, which is based on a dynamic threshold strategy for spiking neurons, and SNNTrack~\cite{zhang2025spiking}, which equips the neurons with a dynamic decay factor. Neither tracker is built entirely from SNNs, and both retain a large number of floating-point multiplications, which prevents their deployment on neuromorphic chips for edge inference.

After SDTrack~\cite{shan2026sdtrack} was proposed, spike-driven trackers built entirely from SNNs became a major line of research. Zhou et al. provided a new optimization scheme named AdaS~\cite{zhouadas}, which helps the training of the tracker converge to the global optimum. Wang et al. proposed a new tracker based on bipolar self-attention~\cite{wang2026bipolar}, which improves the tracking performance considerably. TP-Spikformer~\cite{wei2026tp} obtains a lightweight event-based spike-driven tracker through dynamic computational sparsification and effectively reduces the inference cost. SpikeFET~\cite{yang2026fully} proposes a spatiotemporal regularization (STR) strategy that repairs the similarity degradation of asymmetric features across time steps.

Overall, most recent studies on event-based spike-driven tracking acquire position information with IPL or adopt similar strategies inspired by it. However, why IPL is effective has not been analyzed, and the noise that it introduces in forward and backward propagation has gone unnoticed. Both issues limit the further adoption of IPL and the performance it can deliver. This work provides a mechanistic analysis of IPL and removes this noise through Computation Graph Clipping, which further improves IPL-based trackers without introducing any parameters.

\section{Method}
\subsection{Preliminary}


\subsubsection{Problem Definition}
A standard single object tracker takes one template frame and one search frame as input. Both are obtained by aggregating the event stream over a time interval with an event aggregation method. We denote the template frame by $\mathbf{Z}\in\mathbb{R}^{C\times H_Z\times W_Z}$ and the search frame by $\mathbf{X}\in\mathbb{R}^{C\times H_X\times W_X}$. Given $\mathbf{Z}$, which is cropped around the target, the tracker locates the target in $\mathbf{X}$ and estimates its size. That is, it predicts the bounding box $\hat P=(\hat{c},\hat{s})$ of the target in the coordinate frame of $\mathbf{X}$, where $\hat{c}\in\mathbb{R}^{2}$ is the center and $\hat{s}\in\mathbb{R}^{2}$ contains the width and height. The ground-truth box is denoted by $P^{\ast}=(c^{\ast},s^{\ast})$. The parameters of the tracker are optimized by minimizing the standard tracking loss
\begin{equation}
\label{eq:loss}
\mathcal{L} = {\mathcal{L}_{\operatorname{cls}}} + {\lambda _{\operatorname{iou}}}{\mathcal{L}_{\operatorname{iou}}} + {\lambda _{\mathcal{L}1}}{\mathcal{L}_1},
\end{equation}
where $\mathcal{L}_{\operatorname{cls}}$ is the weighted focal loss applied to the center heatmap, $\mathcal{L}_{\operatorname{iou}}$ and $\mathcal{L}_1$ are the generalized IoU loss and the $\ell_1$ loss applied to the predicted box, respectively, and $\lambda_{\operatorname{iou}}$ and $\lambda_{\mathcal{L}1}$ are fixed weights.

\subsubsection{Spiking Neuron}

The nervous system of the brain is the source of inspiration for neural networks. Its diverse neural dynamics have led researchers to design various types of spiking neurons for SNNs. \textbf{Our method is general and does not rely on any specific neuron type.} We therefore describe spiking neurons with a unified set of dynamical equations:
\begin{align}
\mathbf{U}[t] &= \mathbf{H}[t - 1] + \frac{1}{\tau }(\mathbf{I}[t] - (\mathbf{H}[t - 1] - {\mathbf{U}_{rest}})), \label{eq1} \\
\mathbf{S}[t] &= f(\mathbf{U}[t] - {\mathbf{U}_{thr}}), \label{eq2} \\
\mathbf{H}[t] &= \mathbf{U}[t](1 - \mathbf{S}[t]).  \label{eq3}
\end{align}

When a neuron receives the input $\mathbf{I}[t]$ at time $t$, its membrane potential is updated to $\mathbf{U}[t]$. Eq.~\eqref{eq2} then determines the spike value at time $t$, where $\mathbf{S}[t]=0$ indicates that no spike is emitted and $\mathbf{U}_{thr}$ is the threshold. $\mathbf{H}[t]$ denotes the membrane potential after the spike generation of Eq.~\eqref{eq2}. $\tau$ is the decay factor of the membrane potential, which simulates the decay of the stimulus when the neuron receives no stimulation for an extended period. $\mathbf{U}_{rest}$ denotes the resting potential toward which the membrane potential decays in Eq.~\eqref{eq1}. After a spike is emitted, Eq.~\eqref{eq3} resets the membrane potential to zero.

Mainstream spiking neuron models can be represented by Eqs.~\eqref{eq1}--\eqref{eq3}. For instance, when $f(\cdot)$ is the Heaviside step function, the model represents an IF neuron~\cite{brunel2007lapicque} if $\tau$ equals 1 and an LIF neuron~\cite{maass1997networks} if $\tau$ exceeds 1. When $\mathbf{U}_{rest}=\mathbf{U}_{thr}=0$ and $f(x)={\frac{1}{D} \cdot \operatorname{Clip}(\operatorname{round}(x),0,D)}$, the model becomes the I-LIF neuron~\cite{yao2024scaling}, where $x=\mathbf{U}[t]$ and $D$ is the number of virtual time steps. $\operatorname{Clip}(x, min, max)$ constrains $x$ to $[min, max]$, and $\operatorname{round}(\cdot)$ rounds to the nearest integer. During inference, this model can be converted to binary (0/1) spikes with the spike-ahead principle. In this case, $D$ is folded into the actual number of iterative time steps $T$, giving $T \times D$ time steps in total.

These neurons therefore achieve spike-driven inference, replacing multiply-accumulate (MAC) operations with accumulate (AC) operations and thereby reducing the computational cost substantially. In the following, $\mathcal{SN}(\cdot)$ denotes a layer of spiking neurons governed by Eqs.~\eqref{eq1}--\eqref{eq3}, which maps its input current to the spike output $\mathbf{S}[t]$ position by position. Every convolution and linear projection in the tracker receives the spikes emitted by such a layer. Since all operations other than the neuronal dynamics in Eqs.~\eqref{eq1}--\eqref{eq3} are identical across time steps, the time index $t$ is omitted in the following sections.

\subsubsection{Hierarchical Tracker}

Siamese trackers extract features from $\mathbf{X}$ and $\mathbf{Z}$ with a shared network and model their relation in a separate module. OSTrack~\cite{ye2022joint} and SimTrack~\cite{chen2022backbone} show that a single Vision Transformer can perform feature extraction and relation modeling jointly, so that a convolutional front end is not necessary. Hierarchical architectures such as Swin Transformer~\cite{liu2021swin} and MetaFormer~\cite{yu2023metaformer} nevertheless retain a multi-stage convolutional front end, and their advantages have motivated hierarchical trackers. SDTrack is the representative MetaFormer-style tracker for event-based spike-driven tracking. Its backbone consists of two stages: a convolutional stage $\mathcal{F}^{\mathrm{conv}}_{\theta}$ with $M$ layers and a transformer stage $\mathcal{F}^{\mathrm{attn}}_{\theta}$ built from spiking self-attention (SSA) blocks. The backbone is followed by a tracking head $g(\cdot)$. Layer $m$ of the convolutional stage has kernel size $k_m$ and stride $s_m$. We write $j_m=\prod_{l\le m}s_l$ for the cumulative stride at its output, so that an input of size $H\times W$ yields a feature map of size $(H/j_m)\times(W/j_m)$ at layer $m$. The total stride of the convolutional stage is $j_M$.

\subsubsection{Conventional Design and IPL}
In the conventional pipeline, $\mathbf{X}$ and $\mathbf{Z}$ pass through the convolutional stage separately, with shared or separate weights, are tokenized, receive a positional encoding, and are concatenated before entering the transformer stage. IPL replaces this procedure with a single joint input. The two frames are placed on the diagonal of one tensor, referred to as the joint tensor $\mathbf{U}\in\mathbb{R}^{C\times H_U\times W_U}$ with $H_U=H_X+H_Z$ and $W_U=W_X+W_Z$, which is constructed as
\begin{equation}
\mathbf{U} = \operatorname{IPL}(\mathbf{X},\mathbf{Z}),
\end{equation}
\begin{equation}
\label{eq6}
\operatorname{IPL}(\mathbf{X},\mathbf{Z}) = \left[ {\begin{array}{*{20}{c}}
\mathbf{X}&{{{\rm O}_1}}\\
{{{\rm O}_2}}&\mathbf{Z}
\end{array}} \right],
\end{equation}
where $\mathrm{O}_1\in\mathbb{R}^{C\times H_X\times W_Z}$ and $\mathrm{O}_2\in\mathbb{R}^{C\times H_Z\times W_X}$ are zero blocks. Let $\Omega_X$ and $\Omega_Z$ denote the sets of positions of $\mathbf{U}$ occupied by the search frame and the template frame, respectively. We refer to $\Omega^{+}=\Omega_X\cup\Omega_Z$ as the valid regions, which hold observations. Conversely, the positions of the two zero blocks form the invalid regions $\Omega^{0}$, which hold no observations and exist only to build $\mathbf{U}$.

The joint tensor is processed by the convolutional stage as a single input, which lets the network handle the concatenated frames in the same way as a single-stream input. At the output of the convolutional stage, the two regions are restored from their known locations, tokenized separately, and concatenated before entering the transformer stage. No positional encoding is added at any point, and IPL introduces no parameters. Ideally, the invalid regions would remain inactive throughout the network, so that they add neither information nor computation. Since a zero input produces no spikes, this holds at the input of the network. However, Sec.~\ref{sec:noise} shows that it does not hold in the subsequent layers.

Extensive experiments have shown that IPL outperforms explicit positional encoding~\cite{shan2026sdtrack}, but the underlying reason has not been revealed, which limits further research on IPL-based trackers. The next section provides this analysis.

\subsection{Mechanistic Analysis of IPL}
\label{sec:mechanism}

The convolutional stage $\mathcal{F}^{\mathrm{conv}}_{\theta}(\cdot)$ of SDTrack extracts features from the joint tensor $\mathbf{U}$ obtained by IPL. The template and search frames are then restored to their original spatial structure, tokenized separately, and concatenated. The transformer stage $\mathcal{F}^{\mathrm{attn}}_{\theta}(\cdot)$, which contains spiking self-attention, models the relation between the template and search features. The tracking head $g(\cdot)$ then outputs the predicted position and size of the target. This process can be formalized as the differentiable computation graph
\begin{equation}
\label{eq:graph}
\hat P = g\big(\mathcal{F}_{\theta}(\mathbf{U})\big),
\end{equation}
where $\mathcal{F}_{\theta}$ denotes the backbone, i.e., the convolutional stage $\mathcal{F}^{\mathrm{conv}}_{\theta}$ followed by the restoration and tokenization step and the transformer stage $\mathcal{F}^{\mathrm{attn}}_{\theta}$, and $\theta$ collects all learnable parameters. The graph is optimized with the standard SOT loss in Eq.~\eqref{eq:loss}. In the pair-matching training paradigm of SOT, the template frame is cropped around the target center, so that its center coordinate $P_Z$ corresponds to the position of the target in the template frame. The search frame is cropped from a larger region in which the target may appear, and the ground-truth box $P^{\ast}=(c^{\ast},s^{\ast})$ is defined in the coordinate frame of the search frame. Expressing $P_Z$ in the same coordinate frame, the center of the ground-truth box can be decomposed as
\begin{equation}
\label{eq:decomp}
c^{\ast} = P_Z + \Delta P,
\end{equation}
where $\Delta P$ is the spatial displacement from the center of the template frame to the true position of the target in the search frame. Since the template frame is always cropped around the target, $P_Z$ is a fixed constant, namely the geometric center of the template frame. The variation of $c^{\ast}$ is therefore determined entirely by the relative displacement $\Delta P$. This implies that every term in Eq.~\eqref{eq:loss} is directly governed by $\Delta P$. Consequently, the loss $\mathcal{L}$ in Eq.~\eqref{eq:loss} is essentially a function that is highly sensitive to the relative position between the template and search frames.

With a gradient-based optimizer, the update direction of $\theta$ is determined by $\nabla_{\theta}\mathcal{L}$. By the chain rule, the gradient expands as
\begin{equation}
\label{eq:chain}
\frac{\partial \mathcal{L}}{\partial \theta}
= \frac{\partial \mathcal{L}}{\partial \hat P}\cdot
\frac{\partial \hat P}{\partial \mathcal{F}_{\theta}(\mathbf{U})}\cdot
\frac{\partial \mathcal{F}_{\theta}(\mathbf{U})}{\partial \theta}.
\end{equation}
The first factor encodes the deviation between the predicted and the true position, which, according to Eq.~\eqref{eq:decomp}, is determined by $\Delta P$.

We now examine the structure of the last factor. For a position whose convolutional receptive field overlaps a padded region, the response depends on where that position lies relative to the padding. This boundary effect is the source of the absolute position information encoded by convolutional networks~\cite{islam2020much, kayhan2020translation}. Owing to successive convolutions, the fraction of a feature map that carries such information grows with the depth of the network.

When $\mathbf{U}$ is processed, the receptive field of a convolutional kernel inevitably covers pixels of the template frame, pixels of the search frame, and the zero-padded regions at their boundaries at the same time. Through the convolution operations, this spatial layout causes the relative position information between the two frames to spread through the network layer by layer. Let $\mathcal{N}_{k_m}$ denote the set of kernel offsets of layer $m$, $\mathbf{W}_m[o]$ the kernel weight at offset $o\in\mathcal{N}_{k_m}$, and $\mathcal{R}_m(p)$ the receptive field on $\mathbf{U}$ of output position $p$ of layer $m$. The convolutional stage follows the recursion
\begin{equation}
\label{eq:recursion}
\mathbf{Y}_m(p) = \mathcal{SN}\Big(\sum_{o\in\mathcal{N}_{k_m}} \mathbf{W}_m[o]\,\mathbf{Y}_{m-1}(p+o)\Big),
\end{equation}
for $m=1,\ldots,M$, with $\mathbf{Y}_0=\mathbf{U}$ and batch normalization omitted for brevity. (1) In the first layer, when the kernel moves to within $\lfloor k_1/2\rfloor$ pixels of the boundary between a region and a zero block, $\mathcal{R}_1(p)$ covers valid pixels of that region together with pixels of $\mathrm{O}_1$ and/or $\mathrm{O}_2$. Near the corner at which the two regions meet, it covers pixels of both regions. The output features at these boundary positions implicitly encode the relative spatial layout of the two regions. (2) In the second layer, most kernel positions do not touch both regions at the same time, but their receptive fields cover the boundary features of the first layer. Hence, whenever some $p+o$ is a boundary position of the first layer, $\mathbf{Y}_2(p)$ indirectly acquires position information. (3) After $M$ layers, the receptive-field radius in the coordinates of $\mathbf{U}$ is
\begin{equation}
\label{eq:rf}
r_m = \sum_{l=1}^{m}\Big\lfloor \frac{k_l-1}{2}\Big\rfloor j_{l-1},
\qquad j_0 = 1,
\end{equation}
which reduces to the sum of the kernel half-widths when all strides equal one; each downsampling layer multiplies the growth rate of the receptive field in all subsequent layers by its stride. By translation equivariance, the feature at position $p$ of layer $m$ can depend on the position of $p$ itself only if $\mathcal{R}_m(p)$ contains pixels of a zero block or of the outer padding of the convolution. Otherwise, it depends only on the pixel values within the receptive field. When $r_M$ is comparable to the side lengths of the regions, as is the case for the multi-stage front end of SDTrack, this condition is satisfied throughout the template region and the search region. Even positions inside a region therefore perceive the other region indirectly. Consequently, every position of $\mathbf{Y}_M$ implicitly encodes the relative spatial position of the two regions, which we write as
\begin{equation}
\label{eq:ym}
\mathbf{Y}_M = \mathbf{Y}_M(\mathbf{X},\mathbf{Z},\xi),
\end{equation}
where $\xi$ denotes the layout of $\mathbf{X}$ and $\mathbf{Z}$ in $\mathbf{U}$.

It follows that $\partial \mathcal{F}_{\theta}(\mathbf{U})/\partial\theta$ cannot be decomposed into independent terms that depend only on $\mathbf{X}$ or only on $\mathbf{Z}$; it couples the spatial layouts of the two frames. The first factor of Eq.~\eqref{eq:chain} encodes the prediction error determined by $\Delta P$, the last factor couples the layouts of the two frames through layer-wise propagation, and the chain rule multiplies the two. Therefore, every parameter update is driven by a gradient signal that perceives the relative displacement and the spatial layout at the same time.

\subsection{Noise in the IPL Computation Graph}
\label{sec:noise}
In IPL, $\mathrm{O}_1$ and $\mathrm{O}_2$ are invalid regions introduced to construct the joint tensor $\mathbf{U}$, yet the tracker processes them in the same way as the valid regions. This section examines how this computation introduces noise in convolution, batch normalization, the neuron layer, and attention, and how the noise affects inference and optimization.

\textbf{Convolution.} Eq.~\eqref{eq:recursion} treats all positions of the joint tensor equally. Let $\mathbf{A}_m(q)$ denote the convolutional sum inside the parentheses of Eq.~\eqref{eq:recursion}, i.e., the output of the convolution of layer $m$ at position $q$ before normalization. Let $\Omega^{0}_m$ and $\Omega^{+}_m$ denote the counterparts of $\Omega^{0}$ and $\Omega^{+}$ at the resolution of layer $m$.

In the first layer, the input $\mathbf{U}$ is exactly zero on the invalid regions. Hence, for a position $q\in\Omega^{+}_1$ at the edge of a valid region, the terms of Eq.~\eqref{eq:recursion} with $q+o\in\Omega^{0}$ vanish, and the convolution behaves at this edge exactly as it does at the outer zero padding. This is the boundary effect on which Sec.~\ref{sec:mechanism} relies. For a position $q\in\Omega^{0}_1$ in an invalid region, $\mathbf{A}_1(q)$ is either zero or depends only on the valid content inside $\mathcal{R}_1(q)$. Nevertheless, the shift $\beta_1$ of batch normalization and the subsequent neuron layer cause these positions, which carry no valid information, to emit spikes. The next layer cannot distinguish these spikes from those produced by events. From the second layer on, the inputs on the invalid regions are no longer zero, which has two consequences. First, $\mathbf{A}_m(q)$ is in general nonzero for $q\in\Omega^{0}_m$. Second, the positions at the edge of a valid region no longer receive a fixed zero boundary but rather a response to the neighboring content reflected through the invalid region. The reference against which position is measured is therefore no longer constant.

The invalid regions also take part in backpropagation. Let $\delta_m(q)=\partial\mathcal{L}/\partial\mathbf{A}_m(q)$ denote the gradient backpropagated to position $q$. The weight gradient is
\begin{equation}
\label{eq:wgrad}
\frac{\partial\mathcal{L}}{\partial\mathbf{W}_m[o]}=\sum_{q}\delta_m(q)\,\mathbf{Y}_{m-1}(q+o),
\end{equation}
where $o\in\mathcal{N}_{k_m}$ is the kernel offset in Eq.~\eqref{eq:recursion} and $\mathbf{Y}_{m-1}(q+o)$ is the input to the convolution of layer $m$ that contributes to the output at $q$. Eq.~\eqref{eq:wgrad} sums over all output positions $q$ and all input positions $q+o$, including terms in which $q$ or $q+o$ lies in an invalid region. If the invalid regions did not take part in the computation, the sum would retain only the terms with $q\in\Omega^{+}_m$ and $q+o\in\Omega^{+}_{m-1}$. To express this constraint, let $\mathcal{M}_m$ be the binary validity mask of layer $m$, which equals $1$ on $\Omega^{+}_m$ and $0$ on $\Omega^{0}_m$. The ideal gradient is then $\sum_{q}\mathcal{M}_m(q)\mathcal{M}_{m-1}(q+o)\,\delta^{+}_m(q)\,\mathbf{Y}_{m-1}(q+o)$, where $\delta^{+}_m$ is the backpropagated gradient when the invalid regions do not take part in the computation. Their difference is
\begin{equation}
\label{eq:gdiff}
\sum_{q}\Big[\delta_m(q)-\mathcal{M}_m(q)\mathcal{M}_{m-1}(q+o)\,\delta^{+}_m(q)\Big]\mathbf{Y}_{m-1}(q+o).
\end{equation}
For the terms in which $q$ or $q+o$ lies in an invalid region, the bracket equals $\delta_m(q)$, so the whole term is spurious. Among these terms, those with $q$ in a valid region and $q+o$ in an invalid region correspond to the edge positions that receive input from an invalid region in the forward pass. Those with $q$ in an invalid region have nonzero $\delta_m(q)$ because the leaked activations of the invalid regions feed the valid regions of later layers. For the terms in which both $q$ and $q+o$ lie in valid regions, the bracket equals $\delta_m(q)-\delta^{+}_m(q)$, i.e., the change in the backpropagated gradient itself caused by the contaminated forward pass.

\textbf{Batch normalization.} The statistics of batch normalization are estimated over all positions. Let $\mu_m$ and $\mu^{+}_m$ denote the means of $\mathbf{A}_m$ over $\Omega^{+}_m\cup\Omega^{0}_m$ and over $\Omega^{+}_m$, respectively, and let $\mu^{0}_m$ denote its mean over $\Omega^{0}_m$. Then
\begin{equation}
\label{eq:bnmean}
\mu_m-\mu^{+}_m=\frac{|\Omega^{0}_m|}{|\Omega^{+}_m|+|\Omega^{0}_m|}\big(\mu^{0}_m-\mu^{+}_m\big),
\end{equation}
and an analogous discrepancy holds for the variance. This effect is global: every valid position, including those whose receptive field contains only valid features, is normalized with statistics that depend on the content of the invalid regions. Since the proportion of invalid positions is fixed by the layout and their content is a deterministic function of the input, the perturbation is systematic and persists at inference through the running statistics. In backpropagation, the gradient of batch normalization contains terms averaged over all positions of the layer, so the gradient $\delta_m(q)$ at a valid position is mixed with contributions from invalid positions, which is one source of the difference $\delta_m(q)-\delta^{+}_m(q)$ in Eq.~\eqref{eq:gdiff}. The invalid positions also obtain nonzero gradients in this way and enter the shared weights through Eq.~\eqref{eq:wgrad}.

\textbf{Spiking neuron.} The perturbed normalized values form the input current of the next neuron layer. According to Eqs.~\eqref{eq1}--\eqref{eq3}, a small shift of the membrane potential near the threshold changes the emitted spike and, during training, changes the interval in which the surrogate derivative is nonzero. The invalid regions therefore induce continuous perturbations before spiking and discrete changes of the spike pattern after spiking.

\textbf{Attention.} The valid regions of the output of the convolutional stage are restored and converted into $N_X+N_Z$ tokens, where $N_X=(H_X/j_M)(W_X/j_M)$ and $N_Z=(H_Z/j_M)(W_Z/j_M)$. All of these tokens are valid, but their features already carry the perturbations introduced by convolution, normalization, and the neuron layers described above. Let $\mathbf{q}_a$, $\mathbf{k}_b$, and $\mathbf{v}_b$ denote the query, key, and value vectors of tokens $a$ and $b$ obtained by projecting the token features, and let $\mathbf{y}_a$ denote the output token. The spiking self-attention
\begin{equation}
\label{eq:attn}
\mathbf{Y}^{\mathrm{attn}}=(QK^{\mathsf T})V,\qquad \mathbf{y}_a=\sum_{b}(\mathbf{q}_a\cdot \mathbf{k}_b)\,\mathbf{v}_b,
\end{equation}
expresses each output token as a sum of the values of all tokens weighted by similarity. Consequently, the perturbation carried by any single token becomes a global perturbation after SSA, and the relation modeling between the template and search features deviates accordingly. In backpropagation, the gradient of the value is $\partial\mathcal{L}/\partial \mathbf{v}_b=\sum_{a}(\mathbf{q}_a\cdot \mathbf{k}_b)\,\partial\mathcal{L}/\partial \mathbf{y}_a$, and the gradients of the key and query have the same structure. That is, the gradient of each token is a sum of the output gradients of all tokens weighted by the perturbed similarities. The perturbation therefore affects not only the output of attention but also, in the same way, the gradient propagated back to the convolutional stage.

\textbf{Effect on inference and optimization.} At inference, the features of the valid positions depend on the content of the invalid regions through Eqs.~\eqref{eq:recursion}, \eqref{eq:bnmean}, and \eqref{eq:attn}, which increases the prediction error. During training, the gradient used for the update is
\begin{equation}
\label{eq:gnoise}
g_n=g^{+}_n+\zeta_n
\;\Longrightarrow\;
\langle g_n,g^{+}_n\rangle=\|g^{+}_n\|_2^2+\langle\zeta_n,g^{+}_n\rangle,
\end{equation}
where $g^{+}_n$ is the ideal gradient at iteration $n$, i.e., the gradient when the invalid regions do not take part in the computation. The term $\zeta_n$ collects the differences of Eq.~\eqref{eq:gdiff} over all layers together with the backward perturbations introduced by normalization and attention. The right-hand side follows from taking the inner product of both sides of the first equality with $g^{+}_n$. The perturbation is not zero-mean, because its sign and magnitude are determined by the layout and the input. It therefore does not cancel by averaging over iterations, and any gradient statistics that the optimizer accumulates from the contaminated $g_n$ inherit it. The second term on the right-hand side of Eq.~\eqref{eq:gnoise} measures how the perturbation changes the alignment between the actual gradient and the intended descent direction. Its sign is not fixed a priori. Whenever it is negative, the alignment is reduced. If, in addition, its magnitude exceeds $\|g^{+}_n\|_2^2$, i.e., $\langle\zeta_n,g^{+}_n\rangle<-\|g^{+}_n\|_2^2$, then $\langle g_n,g^{+}_n\rangle<0$ and the update direction is even reversed at that iteration. Since the perturbation does not average out, the optimization path deviates systematically from the intended descent direction, and the parameters converge to an inferior solution. This shows that the original IPL, while providing relative position information, also introduces additional noise that affects both inference and optimization.

\begin{table*}[!htbp]
\centering
\setlength{\tabcolsep}{5pt}      
\begin{tabular}{@{}cccccccccc@{}}
\toprule
\multirow{2}{*}{Methods*}
& \multirow{2}{*}{\begin{tabular}[c]{@{}c@{}}Param.\ (M)\end{tabular}}
& \multirow{2}{*}{\begin{tabular}[c]{@{}c@{}}Spiking \\ Neuron\end{tabular}}
& \multirow{2}{*}{\begin{tabular}[c]{@{}c@{}}Timesteps\\($T\times D$)\end{tabular}}
& \multicolumn{2}{c}{FE108}
& \multicolumn{2}{c}{FELT}
& \multicolumn{2}{c}{VisEvent}
\\
\cmidrule(l){5-10}
& & & &
AUC(\%) & PR(\%)
& AUC(\%) & PR(\%)
& AUC(\%) & PR(\%)
\\
\midrule
STARK~\cite{yan2021learning}
& 28.23 & – & $1\times1$ & 57.4 & 89.2 & 39.3 & 50.8 & 34.1 & 46.8 \\
SimTrack~\cite{chen2022backbone}
& 88.64 & – & $1\times1$ & 56.7 & 88.3 & 36.8 & 47.0 & 34.6 & 47.6 \\
OSTrack\textsubscript{256}~\cite{ye2022joint}
& 92.52 & – & $1\times1$ & 54.6 & 87.1 & 35.9 & 45.5 & 32.7 & 46.4 \\
ARTrack\textsubscript{256}~\cite{wei2023autoregressive}
& 202.56 & – & $1\times1$ & 56.6 & 88.5 & 39.5 & 49.4 & 33.0 & 43.8 \\
SeqTrack-B\textsubscript{256}~\cite{chen2023seqtrack}
& 90.60 & – & $1\times1$ & 53.5 & 85.5 & 33.0 & 42.0 & 28.6 & 43.3 \\
HiT-B~\cite{kang2023exploring}
& 42.22 & – & $1\times1$ & 55.9 & 88.5 & 38.5 & 48.9 & 34.6 & 47.6 \\
GRM~\cite{gao2023generalized}
& 99.83 & – & $1\times1$ & 56.8 & 89.3 & 37.2 & 47.4 & 33.4 & 47.7 \\
HIPTrack~\cite{cai2024hiptrack}
& 120.41 & – & $1\times1$ & 50.8 & 81.0 & 38.2 & 48.9 & 32.1 & 45.2 \\
ODTrack~\cite{zheng2024odtrack}
& 92.83 & – & $1\times1$ & 43.2 & 69.7 & 29.7 & 35.9 & 24.7 & 34.7 \\
SiamRPN~\cite{li2018high}
& – & – & $1\times1$ & – & – & – & – & 24.7 & 38.4 \\
ATOM~\cite{danelljan2019atom}
& – & – & $1\times1$ & – & – & 22.3 & 28.4 & 28.6 & 47.4 \\
DiMP~\cite{bhat2019learning}
& – & – & $1\times1$ & – & – & 37.8 & 48.5 & 31.5 & 44.2 \\
PrDiMP~\cite{danelljan2020probabilistic}
& – & – & $1\times1$ & – & – & 34.9 & 44.5 & 32.2 & 46.9 \\
MixFormer~\cite{cui2022mixformer}
& 37.55 & – & $1\times1$ & – & – & 38.9 & 50.4 & – & – \\
STNet~\cite{zhang2022spiking}
& 20.55 & LIF & $3\times1$ & – & – & – & – & 35.0 & 50.3 \\
SNNTrack~\cite{zhang2025spiking}
& 31.40 & BA-LIF & $5\times1$ & – & – & – & – & 35.4 & 50.4 \\
SpikeET~\cite{yang2026fully}
& 22.36 & I-LIF & $1\times4$ & \textcolor{papergreen}{\textbf{63.9}} &\textcolor{papergreen}{\textbf{93.7}} & – & – &39.9  &54.8       \\
BSA~\cite{wang2026bipolar}
& 19.61 & I-LIF & $1\times4$ & 59.2 & 91.4 & \textcolor{blue}{\textbf{40.9}} & \textcolor{blue}{\textbf{51.8}} & 36.8 & 52.3     \\
MLPixer~\cite{zhang2026unveiling}
& 22.99 & LIF & $4\times1$ & 57.9 & 90.1 & – & – & 34.5 & 48.9        \\
AdaS~\cite{zhouadas}
& 19.61 & I-LIF & $1\times4$ & 60.2 & 92.5 & - & - & 36.3 & 50.5     \\
TP-Spikformer-0.65~\cite{wei2026tp}
& - & I-LIF & $1\times4$ & 59.0 & 91.2 & 39.1 & 50.4 & 35.3 & 49.7
\\
\cmidrule(l){2-10}
\multirow{3}{*}{SDTrack-Tiny~\cite{shan2026sdtrack}}
& \multirow{3}{*}{19.61} & LIF & $4\times1$
& 56.7 & 89.1 & 35.8 & 44.0 & 35.4 & 48.7 \\
& & I-LIF & $2\times2$
& 55.3 & 88.1 & 35.7 & 45.3 & 35.4 & 49.5 \\
& & I-LIF & $1\times4$
& 59.0 & 91.3 & 39.3 & 51.2 & 35.6 & 49.2 \\
\cmidrule(l){2-10}
SDTrack-Base~\cite{shan2026sdtrack}
& 107.26 & I-LIF & $1\times4$
& 59.9 & 91.5 & 40.0 & 51.4 & 37.4 & 51.5 \\
\midrule

\makecell[c]{SDTrack-Tiny$^\ddagger$}
& 22.28 & I-LIF & $1\times4$
&62.8 & 92.8 & 40.5 & 51.4 &39.8 & \textcolor{papergreen}{\textbf{55.0}} \\
\makecell[c]{SDTrack-Base$^\ddagger$}
& 114.40 & I-LIF & $1\times4$
& \textcolor{blue}{\textbf{64.0}} & \textcolor{blue}{\textbf{94.0}} & \textcolor{papergreen}{\textbf{40.8}} & \textcolor{blue}{\textbf{51.8}} & \textcolor{blue}{\textbf{40.5}} &\textcolor{blue}{\textbf{55.2}} \\

\midrule
\makecell[c]{\textbf{SDTrack-Tiny with Mask IPL}}
& 22.28 & I-LIF & $1\times4$
&63.4 &93.6 &40.7 &\textcolor{papergreen}{\textbf{51.7}} &\textcolor{papergreen}{\textbf{40.2}} &\textcolor{blue}{\textbf{55.2}} \\
\cmidrule(l){2-10}
\makecell[c]{\textbf{SDTrack-Base with Mask IPL}}
& 114.40 & I-LIF & $1\times4$
& \textcolor{red}{\textbf{64.4}} & \textcolor{red}{\textbf{94.7}} & \textcolor{red}{\textbf{41.1}} & \textcolor{red}{\textbf{52.2}} &\textcolor{red}{\textbf{40.7}} &\textcolor{red}{\textbf{55.4}} \\
\bottomrule
\end{tabular}
\caption{Comparison with representative ANN-based and SNN-based trackers on three event-based SOT benchmarks. Param.\ is the number of parameters, and $T\times D$ is the number of actual and virtual time steps. $\ddagger$ marks SDTrack retrained with the training strategy and data augmentation in Sec.~\ref{sec:impl}, as the original SDTrack uses no data augmentation; these rows serve as the baseline of Mask IPL. The top three results are shown in \textcolor{red}{\textbf{red}}, \textcolor{blue}{\textbf{blue}}, and \textcolor{papergreen}{\textbf{green}}, respectively.}
\label{main_table}
\end{table*}

\subsection{Mask IPL: Computation Graph Clipping}
\label{sec:iplv2}
The noise originates from the participation of the invalid regions in the computation. As long as the invalid regions are equivalent to the outer zero padding, i.e., their outputs remain zero and they contribute nothing to the gradient, the noise no longer arises. Mask IPL therefore applies the validity mask $\mathcal{M}_m$ to the operations of every layer and clips the computation graph to the subgraph formed by the valid regions.
According to the layout in Eq.~\eqref{eq6}, $\mathcal{M}_m$ equals $1$ on the $(H_X/j_m)\times(W_X/j_m)$ search region at the top left and on the $(H_Z/j_m)\times(W_Z/j_m)$ template region at the bottom right of the $(H_U/j_m)\times(W_U/j_m)$ feature map of layer $m$, and $0$ elsewhere. It is determined solely by the layout and the cumulative stride and contains no learnable parameters.

Specifically, when the tracker adopts Mask IPL, the $m$-th Conv-BN-$\mathcal{SN}$ block computes
\begin{equation}
\label{eq:clip}
\widetilde{\mathbf{Y}}_m=\mathcal{SN}\big(\operatorname{BN}_{\mathcal{M}}(\mathcal{M}_m\odot\mathbf{A}_m)\big),
\end{equation}
where $\odot$ denotes element-wise multiplication broadcast over channels, and $\operatorname{BN}_{\mathcal{M}}$ normalizes with the statistics of the valid regions only and outputs zero on $\Omega^{0}_m$. Its mean is
\begin{equation}
\label{eq:maskmean}
\widetilde\mu_m=\frac{1}{|\Omega^{+}_m|}\sum_{q\in\Omega^{+}_m}\mathbf{A}_m(q),
\end{equation}
its variance is defined analogously, and the running statistics are accumulated from both. Since $\operatorname{BN}_{\mathcal{M}}$ outputs zero on $\Omega^{0}_m$ and a neuron that receives zero input emits no spike, the output of Eq.~\eqref{eq:clip} is zero on the invalid regions. For a block with a residual connection, the quantity added to the output of Eq.~\eqref{eq:clip} is the input $\widetilde{\mathbf{Y}}_{m-1}$ of that block. The sum remains zero on the invalid regions as long as this input is zero there. By induction from $\widetilde{\mathbf{Y}}_0=\mathbf{U}$, $(1-\mathcal{M}_m)\odot\widetilde{\mathbf{Y}}_m=0$ therefore holds for every layer. The input of every convolution is thus exactly zero on the invalid regions, as it is on the outer zero padding. The boundary mechanism of Sec.~\ref{sec:mechanism} is therefore preserved, whereas the perturbations produced by the invalid regions in Sec.~\ref{sec:noise} no longer arise. In backpropagation, since $\partial(\mathcal{M}_m\odot\mathbf{A}_m)/\partial\mathbf{A}_m=\mathcal{M}_m$, the positions in the invalid regions receive no gradient, and since the features on the invalid regions are zero, the weight gradient takes the form
\begin{equation}
\label{eq:clipgrad}
\frac{\partial\mathcal{L}}{\partial\mathbf{W}_m[o]}=\sum_{q}\mathcal{M}_m(q)\mathcal{M}_{m-1}(q+o)\,\delta_m(q)\,\widetilde{\mathbf{Y}}_{m-1}(q+o),
\end{equation}
and the input gradient likewise propagates only along connections with both ends in the valid regions. Eq.~\eqref{eq:clipgrad} has the same form as the ideal gradient in Sec.~\ref{sec:noise}, and the normalization statistics likewise come from the valid regions only. Hence, the actual gradient is the ideal gradient, and $\zeta_n$ in Eq.~\eqref{eq:gnoise} vanishes. The update direction coincides exactly with the intended descent direction, and the deviation of the optimization path described in Sec.~\ref{sec:noise} is eliminated. Eq.~\eqref{eq:clip} applies to all convolutions, linear projections, and batch normalizations of both the convolutional and the transformer stages. Under this rule, the restored tokens no longer carry the perturbations described in Sec.~\ref{sec:noise}, and the projections and MLPs of the transformer stage introduce no new perturbations. The attention in Eq.~\eqref{eq:attn} therefore requires no additional treatment. Mask IPL thereby preserves the boundary mechanism through which the convolutional stage learns the relative position between the template and search frames, introduces no parameters, and makes the actual gradient coincide with the ideal gradient.

\section{Experiments}

\subsection{Implementation Details}
\label{sec:impl}

The networks used in this study are identical to those of SDTrack, except for two modifications. First, a depthwise separable convolution block is inserted before each attention module. Second, a type coding is introduced, which assigns a separate learnable encoding to the template frame and to the search frame. We first pre-train the backbone on ImageNet-1K with the same pre-training strategy as SDTrack and then fine-tune it on the pair-matching task with 60,000 sample pairs per epoch. Training uses the AdaNSD optimizer~\cite{zhouadas} with cosine learning rate decay. On FE108~\cite{zhang2021object} and VisEvent~\cite{wang2023visevent}, the tracker is trained for 50 epochs with the Voxel event representation and standard data augmentation; on FELT~\cite{wang2024long}, it is trained for 300 epochs with the GTP event representation and without any data augmentation. All experiments run on four H100 GPUs with CUDA~13.0 and PyTorch~2.14.

\subsection{Main Results}


We compare the proposed method with representative ANN-based and SNN-based trackers on FE108, FELT, and VisEvent in Tab.~\ref{main_table}. For a fair comparison, we reproduce SDTrack with the same training strategy and data augmentation as Mask IPL (Sec.~\ref{sec:impl}). The original SDTrack, in contrast, is trained without data augmentation. The reproduced models are marked with $\ddagger$ and serve as the baselines of Mask IPL. In our experiments, data augmentation combined with the Voxel event representation performs considerably better on FE108 and VisEvent. On FELT, however, it performs clearly worse. These results will be provided in a later version of this paper, possibly with a detailed analysis.

Overall, we find that inserting a convolution block before each attention module noticeably improves tracking performance when data augmentation is applied. Following SpikeFET~\cite{yang2026fully}, we adopt this design (Sec.~\ref{sec:impl}). Consequently, both the reproduced SDTrack and the trackers with Mask IPL contain more parameters than the original SDTrack. The increase is 2.67M at the Tiny scale and 7.14M at the Base scale. The type coding described in Sec.~\ref{sec:impl} also contributes to this increase. However, it adds only a few thousand parameters, which is negligible. Mask IPL itself introduces no parameters. Each tracker with Mask IPL therefore has the same number of parameters as its reproduced baseline.

Replacing IPL with Mask IPL improves the reproduced SDTrack-Tiny on all three benchmarks. On FE108, FELT, and VisEvent, the AUC increases by 0.6, 0.2, and 0.4 percentage points, respectively. The PR increases by 0.8, 0.3, and 0.2 percentage points, respectively. This improvement also extends to the Base scale on FE108, FELT, and VisEvent.

More notably, Mask IPL can reduce the amount of computation on any hardware platform. This requires only rewriting or modifying the convolution operators. Mask IPL clips the computation graph to the subgraph formed by the valid regions (Sec.~\ref{sec:iplv2}). The modified operators can therefore skip the invalid regions. In contrast, the corresponding reduction for IPL is available only on neuromorphic chips. Even there, it is limited by the spread of the noise analyzed in Sec.~\ref{sec:noise}. Because of this spread, the invalid regions emit spikes after the first layer and still trigger computation. The modified operators will be released in a later version.

\section{Conclusion}

This paper analyzes Intrinsic Position Learning (IPL), a parameter-free way of acquiring position information in event-based spike-driven tracking, and explains its effectiveness. The zero blocks of the joint tensor are equivalent to zero padding for convolution, and the resulting boundary effect spreads layer by layer through the multi-stage convolution. Every parameter update is therefore driven by a gradient that perceives the relative displacement between the template and search frames. A positional encoding added after the convolutional stage cannot provide this information.

The same analysis shows that the invalid regions of the joint tensor take part in convolution, batch normalization, the neuron layers, and attention. They thereby introduce feature noise in forward propagation and gradient noise in backward propagation. The former increases the inference error, and the latter causes the optimization path to deviate from the ideal descent direction. We accordingly propose Computation Graph Clipping. It applies a validity mask determined by the layout to the operations of every layer, so that the invalid regions become equivalent to zero padding in both forward and backward propagation. The boundary mechanism is preserved, the noise is eliminated, and the actual gradient coincides with the ideal gradient, all without any additional parameters.

The resulting Mask IPL improves both the Tiny-scale and the Base-scale trackers on FE108, FELT, and VisEvent. These results indicate that efficiently modeling the relative position between the template and search frames is one of the key factors for event-based SOT. The analysis also offers a basis for the further design of IPL-based trackers.

\section{Limitations}


This manuscript is released in its current form so that Mask IPL can be made available as early as possible as a more robust alternative for acquiring position information in event-based spike-driven tracking. Owing to the limited preparation time, some of the content is not yet presented with full rigor, and we welcome comments from readers. Most of the material will be reorganized in the formal version, whose structure and presentation may differ considerably.


\bibliographystyle{IEEEtran}
\bibliography{ref}

\end{document}